\documentclass[runningheads]{llncs}
\usepackage{graphicx}
\usepackage{caption}
\usepackage{subcaption}
\usepackage{multirow}
\usepackage{hyperref}
\begin{document}
%
% \title{No Budget? Don’t Flex! Cost Consideration when Planning to Adopt NLP for Your Business}
\title{Costs to Consider in Adopting NLP \\ for Your Business}
\titlerunning{Costs to Consider in Adopting NLP for Your Business}
% If title is too long set an abbreviated paper title here
%

\author{
Made Nindyatama Nityasya\inst{1}\orcidID{0000-0002-5570-4676} \and \\
Haryo Akbarianto Wibowo\inst{1}\orcidID{0000-0001-7457-2165} \and \\
Radityo Eko Prasojo\inst{1,2}\orcidID{0000-0002-5148-7299} \and \\
Alham Fikri Aji\inst{1}\orcidID{0000-0001-7665-0673}
} 
\authorrunning{M.N. Nityasya et al.}
\institute{
Kata.ai Research Team, Jakarta, Indonesia\\ 
\email{\{made,haryo,ridho,aji\}@kata.ai}
\and
Faculty of Computer Science, Universitas Indonesia
%\and
%School of Informatics, University of Edinburgh
}

\maketitle

\begin{abstract}
Recent advances in Natural Language Processing (NLP) have largely pushed deep transformer-based models as the go-to state-of-the-art technique without much regard to the production and utilization cost.
Companies planning to adopt these methods into their business face difficulties because of the lack of machine, data, and human resources to build them.
We compare both the performance and the cost of classical learning algorithms to the latest ones in common sequence and text labeling tasks.
In our industrial datasets, we find that classical models often perform on par with deep neural ones despite the lower cost. 
We show the trade-off between performance gain and the cost across the models to give more insights for AI-pivoting business.
% We argue that under many circumstances the smaller and lighter models fit better for AI-pivoting businesses and that w 
Further, we call for more research into low-cost models, especially for under-resourced languages.


\end{abstract}

\section{Introduction}

Research on benchmarking learning algorithms for NLP tasks~\cite{1904.08067,2004.03705,yadav-bethard-2018-survey,guntara-etal-2020-benchmarking} have largely focused on the quality of the models by some accuracy metrics such as the F1 score.
The costs which include processing time, memory resources, computing power, and human expertise that is needed to train the models and be utilized for prediction are often ignored.  
 
As NLP is getting popular to be implemented across industries,
one of the biggest hurdles in this early adoption is determining which methods to use. Companies want to provide the best model but often had struggled with resources to build it~\cite{magoula2020ai}.
This is especially true for companies operating within a large emerging market with under-resourced languages, which often means the lack of human expertise, data limitation, short experiment time, and budget limitation. For companies already serving a large number of users, they also need to consider how the model scales and keeps prediction time fast even without expensive GPU servers.

In this paper, we focus on the NLP industry landscape of the large emerging market of Indonesia, where highly demanded products are based on text classification and sequence labeling~\cite{ruliputra-2019-enterprise}. These two tasks has under-resourced annotated data in Indonesian language~\cite{wilie2020indonlu}. We run experiments 
over a number of datasets using several learning algorithms.
We report F1-scores, training time, resulting model's size, and average prediction time. We discuss how our experiment results have influenced our business decision
% as one of the industry players in Indonesia 
and how it can help other companies in adapting NLP technologies fast and more effectively.

\section{Indonesian NLP Landscape}
\begin{figure}
    \vspace{-30px}
    \centering
    \includegraphics[width=160pt]{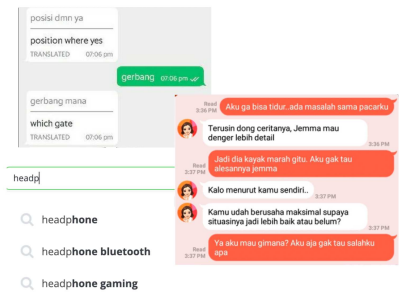}
    \vspace{-10px}
    \caption{Several NLP applications in production in Indonesia: machine translation in chat (top-left), chatbot (right), product search engine (bottom-left)} \label{fig:nlp-app}
    \vspace{-10px}
\end{figure}

Over the last few years, NLP has gained popularity across industries in Indonesia. From creating chatbots, analyzing customer's reviews, machine translation for chat, and improving product search engine (see Figure~\ref{fig:nlp-app}), many companies progressively adopt NLP into their business.
Amongst all, text analytics and chatbots are the most popular NLP products~\cite{bahja2020natural}, which utilize text classification and sequence labeling as their base tasks.

Indonesia's market shows great potential for AI and NLP adoption~\cite{chitturu2017artificial}.
Unfortunately, Indonesian language is still under-researched on many NLP tasks. It possesses multiple challenges, ranging from its ever-changing, ubiquitous, vastly-varying colloquialisms~\cite{wibowo2020semi} to its lack of labeled data in many tasks and domains. Common academic literature suggests the use of multilingual models or transfer learning~\cite{howard-ruder-2018-universal,ruder2019transfer} to combat low-resourceness, but the large model size makes them expensive to train and to run. Even if we use knowledge distillation~\cite{hinton2015distilling} to make a smaller and faster model for prediction, it requires a teacher system which use %is usually an ensemble of multiple 
larger neural models~\cite{kim2019research}. Thus, creating a lightweight model often increases the training cost.

Another challenge for Indonesian companies is the use of GPUs for training and prediction, which is arguably a basic requirement for the latest NLP methods, yet only a limited number of institutions has utilized GPU servers.\footnote{Indonesian AI National Strategy; Source in Indonesian: \url{https://ai-innovation.id/server/static/ebook/stranas-ka.pdf}, Section 5}
In addition to that, deep learning has just gained its popularity in recent years, coupled with a slow adoption of deep learning into the standard study curriculum by Indonesian government, making the experts in this field is still scarce on the industry level.\footnote{Indonesian AI National Strategy, Section 4}

Knowing the constraints, we explore learning algorithms from the classic to the latest ones. By benchmarking the performance and costs, we provide more insight and perspective for early-adoption of NLP in businesses.
% Knowing the constraints, we explore learning algorithms from the classical ones to the latest ones in the hope of providing more insight and perspective for early-adopting NLP in business.

\section{Learning Algorithms}
\label{sec:learning-alg}
Here we describe several algorithms that we use in this paper from the classical methods to the latest deep neural models.

\subsection{Classical ML}

Logistic regression (LR) and support vector machine (SVM) are often used to train models for text classification. Before the advent of deep learning, SVM models often top the chart in text classification tasks~\cite{manevitz2001one}. These two methods are also popular and within the repertoire of a typical Indonesian data scientists and engineers.

Conditional Random Field (CRF) is a common technique for sequence labeling. Even in deep neural methods, CRF is often used as the last layer to improve the performance~\cite{ma-hovy-2016-end,CHEN2017221}. In this work, we only use basic features, such as orthography, prefix, suffix, bigram, and trigram, without external knowledge.

\subsection{Bi-LSTM}
\label{sec:bi-lstm}

\begin{figure}[htp]
    \centering
    \includegraphics[width=200pt]{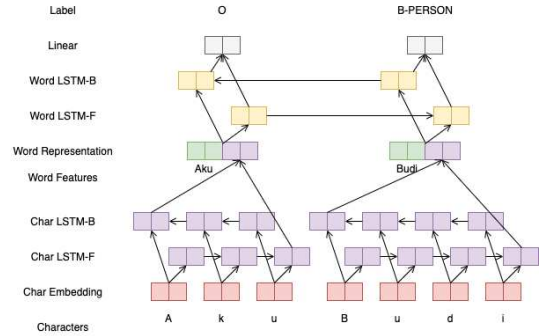}
    \caption{Character-Word level BI-LSTM. 'Aku Budi' means 'I am Budi'.}
    \label{fig:char-word-bilstm}
    \vspace{-10px}
\end{figure}

Bidirectional-LSTMs (Bi-LSTM) are able to capture information from long sequences in forward and backward direction. It shows good results~\cite{huang2015bidirectional,CHEN2017221} and until recently was the go-to state-of-the-art technique for sequential data like text. We stack two Bi-LSTM layers to capture the information from both character and word level as shown in Figure~\ref{fig:char-word-bilstm}. The result from the character level is concatenated with the word embedding before getting passed into another Bi-LSTM layer. For text classification, we concatenate the output from the forward and the backward word level layer, before passing it to a dense layer to get the result.

\subsection{Convolutional Neural Network}

Although it has roots in computer vision, Convolutional Neural Network (CNN) has been shown to perform well for text classification~\cite{kim-2014-convolutional,johnson-zhang-2015-effective} via 1-dimensional convolution to capture sequence of words. We use Kim's~\cite{kim-2014-convolutional} architecture for text classification with some adjustments. Instead of word2vec~\cite{41224}, we use FastText~\cite{joulin2016bag} to handle out-of-vocabulary (OOV) words. For sequence labeling, we modify the previous character level representation (Subsection~\ref{sec:bi-lstm}, Figure~\ref{fig:char-word-bilstm}) to CNN before concatenating it to the word-level embedding and feeding it into the Bi-LSTM layer.

\subsection{Transformers}

The Transformer has become the latest state-of-the-art method in many NLP tasks as it has shown to outperform other neural models like RNN or LSTM~\cite{2004.03705}.
With sufficient computing power, it can run faster (relative to the RNNs) because of its ability to run in parallel~\cite{vaswani2017attention}. The Transformer also gave rise to pre-trained language models such as BERT~\cite{devlin-etal-2019-bert} and ALBERT~\cite{lan2020albert}, which is a lighter version of BERT with lower memory consumption.

Here, we use inductive transfer learning to extract knowledge from existing language models and fine-tune it to the downstream tasks. This method has shown the biggest improvement and widely used in many NLP applications~\cite{ruder2019transfer}.

\section{Algorithm benchmark}
In this section, we describe how we benchmark the algorithms in Section~\ref{sec:learning-alg}.

\subsection{Tasks and Datasets}

We run our experiments across multiple actual industry's datasets. However, some of the data used in our experiments are private, thus cannot be published. Nevertheless, we provide data description and statistic (see Table~\ref{eda-data}) to give a general view of what the data is about. 

\begin{table}
    \centering
    \caption{Data statistics. Train: total training data. Dev: total validation data. Test: total testing data. $N$: total data. $c$: total label. $d$: average data per class for classification. $l$: average sentence length. $V$: vocab size. $ts$: average sentence length for 100 sample prediction data.}
    \begin{tabular}{l|rrrr|rrrr}
        \hline
        \textbf{} & \textbf{Smltk} & \textbf{Health} & \textbf{Telco} & \textbf{Sent} & \textbf{EntK} & \textbf{POS} & \textbf{TermA} & \textbf{Prod} \\
        \hline
        {Train} & {11134} & {57938} & {11520} & {28717} & {10955} & {3000} & {7222} & {1365} \\
        {Dev} & {1280} & {6894} & {1440} & {3191} & {1250} & {1000} & {802} & {854} \\
        {Test} & {1272} & {6897} & {1440} & {4748} & {1372} & {1000} & {2006} & {853} \\
        {$N$} & {13686} & {71729} & {14400} & {36656} & {13577} & {5000} & {10030} & {3072} \\
        {$c$} & {96} & {5} & {144} & {2} & {14} & {3} & {23} & {69} \\
        {$d$} & {142.51} & {14345.8} & {100} & {18328} & {-} & {-} & {-} & {-} \\
        {$l$} & {5.08} & {8.56} & {5.27} & {15.6} & {12.36} & {15.72} & {26.11} & {9.61} \\
        {$V$} & {2878} & {16892} & {3357} & {22896} & {18004} & {5211} & {15624} & {5655} \\
        {$ts$} & {4.88} & {9.06} & {5.16} & {19.14} & {12.13} & {15.81} & {26.47} & {9.93} \\
        \hline
        \end{tabular}
    \label{eda-data}
\vspace{-10px}
\end{table}

\noindent{\textbf{Text Classification Task}}

\begin{itemize}
    \item \textbf{Smltk}
Bot intent classification for small talk (e.g. greetings, joking, etc.). The language is informal and the labels are imbalanced. 
    \item \textbf{Health}
Text classification for conversation between doctors and patients. The data is semi-formal and grouped into five labels: patient's complaint, patient's action, doctor's diagnosis, doctor's recommendation, and other.
    \item \textbf{Telco}
Intent classification for a telecommunication’s bot. It contains semi-formal question and instruction with a balanced data across labels.
    \item \textbf{Sent}\footnote{\url{https://www.kaggle.com/grikomsn/lazada-indonesian-reviews}}
Sentiment analysis data about product reviews from an e-commerce platform. The data was annotated based on user's rating.
\end{itemize}

\noindent{\textbf{Sequence Labeling Task}}

\begin{itemize}
    \item \textbf{EntK}
Extended NER with 14 different labels including person, location, email, phone, datetime, number, currency and 5 different units. This is our internal dataset that was manually gathered and annotated.
    \item \textbf{POS}\footnote{\url{https://github.com/kmkurn/id-pos-tagging}}
POS Tagging dataset from the PAN Localization Project~\cite{dinakaramani2014designing}. We use one of the data splits that was done by~\cite{kurniawan2018}.
    \item \textbf{TermA}\footnote{\url{https://github.com/jordhy97/final\_project}}
A semi-formal review data from AiryRoom, a hotel aggregator platform. The data is annotated into aspects and their sentiment~\cite{fernando2019aspect}.
    \item \textbf{Prod}\footnote{\url{https://github.com/derhif/enamex-center}} 
This dataset contains product title with its annotated attributes from several e-commerce websites in Indonesia~\cite{articlerifat}. 

\end{itemize}

\subsection{Experiment Setup}

For training we use a single GPU of Tesla T4 15GB. Whereas for prediction, we compare the same GPU machine with a CPU Intel(R) Xeon(R) CPU @ 2.20GHz (4 cores) with memory of 26.75 GB.

We use TF-IDF weighted $n$-grams ($n =$ 1, 2) as the word vector in our classic methods. For the FastText, we use the pretrained Indonesian word vector\footnote{\url{https://fasttext.cc/docs/en/crawl-vectors.html}} with 300-dimension then fine-tune it on our training dataset.
For pre-trained language models, we compare two base models. The first one is a multilingual BERT (mBERT) base model\footnote{\url{https://huggingface.co/bert-base-multilingual-cased}}~\cite{devlin-etal-2019-bert} which contains 104 different languages, including Indonesian language. 
This model often used as base pre-trained model for non-English dataset. 
Secondly, we use IndoNLU's~\cite{wilie2020indonlu} lite model\footnote{\url{https://huggingface.co/indobenchmark/indobert-lite-base-p1}}, an ALBERT~\cite{lan2020albert} base model trained using Indonesian dataset.

We use AdamW Optimizer~\cite{loshchilov2017decoupled} with the learning rate of 1e-3, 1e-4, 1e-5. The batch size is set to 16, following the recommended hyperparameter settings~\cite{wilie2020indonlu,devlin-etal-2019-bert}, for all datasets and methods to ensure fairness in the experiment. 
We use early stopping based on 3 consecutive validation loss.

\subsection{Evaluation Metrics} 

Other than the model quality, we also observe training time, size of the model, as well as its loading and prediction time.

\noindent {\textbf{F1 Score:}} F1-macro is used as our evaluation metric to average over classes. This metric is used for both binary and multi-label classification, also for the sequence labeling. 

\noindent {\textbf{Training Resources:}} When training the neural models, we use one GPU and track the total training time. Because hyperparameter tuning is costly, we run the training once using the recommended settings. We also track every saved file that represents the model to calculate the total model size. In the case of a company where the business provides a Platform-as-a-Service (PaaS), 
this size corresponds to the amount of storage and memory needed to load the model, and it scales according to the number of users/clients.

\noindent{\textbf{Load Time:}} Loading the model into memory is a prerequisite before it can be used. In the case where we have hundreds of models with a limited machine resource, it is impossible to always host all models, especially if the models are not used often. Periodically, the model would be removed from memory and be rebuild when it is needed. Knowing that loading time becomes information that needs to be taken into account.

\noindent{\textbf{Prediction Time:}} We compare the prediction time between using one CPU and one GPU. For each dataset, we take
100 stratified random samples based on token’s length. We run prediction one-by-one for 100 samples then sum its prediction time. To heighten the accuracy of our experiment, we rerun the prediction using pytest-benchmark,\footnote{\url{https://pypi.org/project/pytest-benchmark}} which automatically minimizes outliers, for 100 rounds.
For the classic algorithms, we only run on the CPU.

\section{Experiment Results}
\begin{table*}[ht!]
% \vspace{-20px}

\centering
\caption{\label{result-f1-traintime}
F1: macro F1-score from test set. T: training time (seconds). E: total epochs. $^\dag$\texttt{mBERT} for Sent dataset encounter CUDA out-of-memory.
}

\begin{tabular}{@{}lrrrrrrrrrrrr@{}}

\hline
\textbf{Method} & {\textbf{F1}} & \textbf{T} & \textbf{(E)}  & \textbf{F1} & \textbf{T} & \textbf{(E)}  & \textbf{F1} & \textbf{T} & \textbf{(E)} & \textbf{F1} & \textbf{T} & \textbf{(E)} \\
\hline
{} & \multicolumn{3}{c@{\hspace{2.5em}}}{\textbf{Smltk}} & \multicolumn{3}{c@{\hspace{2.5em}}}{\textbf{Health}} & \multicolumn{3}{c@{\hspace{2.5em}}}{\textbf{Telco}} & \multicolumn{3}{c}{\textbf{Sent}} \\
\hline
{\texttt{LR}} & {0.763} & {23.23} & -- & {0.697} & {21.15} & -- & {0.857} & {28.21} & -- & {0.793} & {9.10} & -- \\
{\texttt{SVM}} & {0.902} & {2.16} & -- & {0.730} & {8.96} & -- & {0.904} & {2.65} & -- & {0.850} & {6.76} & -- \\
{\texttt{Bi-LSTM}} & {0.787} & {186.16} & {(39)} & {0.700} & {894.71} & {(16)} & {0.826} & {374.62} & {(41)} & {0.806} & {223.11} & {(9)} \\
{\texttt{CNN}} & {0.899} & {134.06} & {(31)} & {0.724} & {222.66} & {(6)} & {0.902} & {201.07} & {(32)} & {0.851} & {142.59} & {(7)} \\
{\texttt{mBERT}} & {0.925} & {1459.60} & {(14)} & {0.780} & {4345.80} & {(5)} & {0.919} & {1676.55} & {(15)} & \multicolumn{3}{c}{NA$^\dag$} \\
{\texttt{IndoNLU}} & \textbf{0.940} & {518.64} & {(13)} & \textbf{0.807} & {1499.40} & {(3)} & \textbf{0.941} & {946.49} & {(22)} & \textbf{0.868} & {1385.34} & {(3)} \\
\hline
{} & \multicolumn{3}{c@{\hspace{2.5em}}}{\textbf{EntK}} & \multicolumn{3}{c@{\hspace{2.5em}}}{\textbf{POS}} & \multicolumn{3}{c@{\hspace{2.5em}}}{\textbf{TermA}} & \multicolumn{3}{c}{\textbf{Prod}} \\
\hline
{\texttt{CRF}} & {0.867} & {55.88} & -- & {0.958} & {67.84} & -- & {0.884} & {5.45} & -- & \textbf{0.778} & {85.44} & -- \\
{\texttt{Bi-LSTM}} & {0.850} & {1065.97} & {(11)} & {0.945} & {942.61} & {(13)} & {0.817} & {338.46} & {(11)} & {0.705} & {591.51} & {(38)} \\
{\texttt{CNN}} & {0.848} & {294.89} & {(10)} & {0.942} & {257.31} & {(11)} & {0.817} & {104.25} & {(9)} & {0.682} & {105.26} & {(30)} \\
{\texttt{mBERT}} & {0.891} & {1154.19} &{(6)} & \textbf{0.965} & {1394.94} & {(9)} & {0.871} & {297.24} & {(4)} & {0.666} & {393.55} & {(16)} \\
{\texttt{IndoNLU}} & \textbf{0.899} & {662.21} & {(6)} & {0.957} & {715.75} & {(8)} & \textbf{0.890} & {193.23} & {(5)} & {0.711} & {255.87} & {(23)} \\
\hline
\end{tabular}
\vspace{-10px}
\end{table*}

The ALBERT-based model of IndoNLU achieves the best results for almost all experiments (see Table~\ref{result-f1-traintime}), 
even when we portion the training data to see whether data size affects the results (see Table~\ref{res-percentage}).
Interestingly, the best classical approaches yield competitive results that are on average only around 0.03 lower than IndoNLU in terms of F1 score. On EntK data, we also found that the F1 score difference between the IndoNLU and CRF decrease as the size of training data increase.

\begin{table}[]
\vspace{-20px}
\caption{F1 Score with portions (in percentage) of training data}
\begin{subtable}[h]{0.53\textwidth}
\centering
\begin{tabular}{l|rrrrr}
\hline
& \textbf{2.5} & \textbf{25} & \textbf{50} & \textbf{75} & \textbf{100} \\ 
\hline
train size            & 1446  & 14483 & 28966 & 43455 & 57938  \\ 
\hline
{\texttt{SVM}}        & 0.582 & 0.689 & 0.713 & 0.726 & 0.730 \\
{\texttt{CNN}}        & 0.588 & 0.688 & 0.712 & 0.730 & 0.733 \\
{\texttt{IndoNLU}}    & 0.626 & 0.744 & 0.783 & 0.798 & 0.803 \\
\hline
\end{tabular}
\caption{classification task using Health data}
\end{subtable}
\hfill
\begin{subtable}[h]{0.49\textwidth}
\centering
\begin{tabular}{l|rrrr}
\hline
& \textbf{25} & \textbf{50} & \textbf{75} & \textbf{100} \\
\hline
train size          & 2708  & 5416  & 8247  & 10955  \\ 
\hline
{\texttt{CRF}}      & 0.787 & 0.831 & 0.857 & 0.868 \\
{\texttt{Bi-LSTM}}  & 0.675 & 0.734 & 0.793 & 0.833 \\
{\texttt{IndoNLU}}  & 0.854 & 0.879 & 0.896 & 0.899 \\
\hline
\end{tabular}

\caption{labeling task using EntK data}
\label{res-percentage-2}
\end{subtable}
\label{res-percentage}
\vspace{-25px}
\end{table}

A unique case occurs with the Prod dataset where the CRF slightly outperforms Transformer. The Prod dataset has a large number of labels (69) compared to its small amount of training instances (1365) which might not be sufficient for Transformers. Also, despite using GPU, when the base model and the dataset are large, we can still encounter memory limitation issue like we did when training BERT using the largest dataset (Sent). Training failure means that it has to be repeated in some other way, which means additional costs. The memory error might be addressed by procuring bigger, better GPUs, which is another cost.

\begin{table}[]
% \vspace{-20px}
\centering
\caption{Resulting model size (MB)}
\begin{subtable}[h]{0.49\textwidth}
\centering
\begin{tabular}{lrrrr}
% & & & & \\
\hline
\textbf{Method} & \textbf{Smltk} & \textbf{Health} & \textbf{Telco} & \textbf{Sent} \\
\hline
{\texttt{LR}} & {10.75} & {8.67} & {22.41} & {6.79} \\
{\texttt{SVM}} & {10.75} & {8.67} & {22.41} & {6.79} \\
{\texttt{Bi-LSTM}} & {3.91} & {13.50} & {4.39} & {17.39} \\
{\texttt{CNN}} & {17.30} & {76.51} & {19.67} & {100.39} \\
{\texttt{mBERT}} & {2135.26} & {2134.42} & {2135.71} & {NA} \\
{\texttt{IndoNLU}} & {141.13} & {140.28} & {141.57} & {140.26} \\
\hline
\end{tabular}
\caption{classification task}
\end{subtable}
\hfill
\begin{subtable}[h]{0.49\textwidth}
\centering
\begin{tabular}{lrrrr}
%& & & & \\
\hline
\textbf{Method} & \textbf{EntK} & \textbf{POS} & \textbf{TermA} & \textbf{Prod} \\
\hline
{\texttt{CRF}} & {3.09} & {2.70} & {1.00} & {3.53} \\
{\texttt{Bi-LSTM}} & {11.60} & {9.86} & {3.95} & {4.72} \\
{\texttt{CNN}} & {14.36} & {12.15} & {4.66} & {5.63} \\
{\texttt{mBERT}} & {2129.88} & {2129.84} & {2129.68} & {2130.78} \\
{\texttt{IndoNLU}} & {135.74} & {135.71} & {135.54} & {136.64} \\
\hline \\
\end{tabular}

\caption{labeling task}
\end{subtable}
\label{model-size}
\vspace{-20px}
\end{table}

The result shows a small difference in F1 score between classical and neural models despite being much more resource-efficient. They train in seconds (on CPU) and hundreds of times faster than the Transformer which requires GPU. Similarly, the prediction can be done on the CPU and is about 10x faster than the prediction time of the Transformer-based models using the GPU (see Table~\ref{model-load-infer}). 

In terms of model size, the classical methods are heavily affected by the amount of vocab and the labels whereas the neural model's size depends on the complexity of the model's architecture (see Table~\ref{model-size}). The variance of model size in Transformer based model are smaller.
For early adaptation, it is expensive to have multiple customized BERT model as 1 model reach up to 2 GB of storage, especially if we want to have model versioning. The lighter version that we use with IndoNLU shows better feasibility to be deployed into production.

\begin{table}[]
\vspace{-20px}
\caption{Avg. loading and prediction time (seconds)}
\centering
\begin{subtable}[h]{0.50\textwidth}
\begin{tabular}{lrrrr}
\hline
\multirow{2}{*}{\textbf{Method}} & \multicolumn{2}{c}{\textbf{load}} & \multicolumn{2}{c}{\textbf{infer}} \\
    & \textbf{CPU} & \textbf{GPU} & \textbf{CPU} & \textbf{GPU} \\
\hline
\multicolumn{5}{c}{\textbf{Classification}} \\
\hline
{\texttt{LR}}      & {0.025}  & {--}     & {0.580} & {--} \\
{\texttt{SVM}}     & {0.027}  & {--}     & {0.153} & {--} \\
{\texttt{Bi-LSTM}} & {3.963}  & {4.044}  & {0.166} & {0.123}  \\
{\texttt{CNN}}     & {4.707}  & {4.613}  & {0.154} & {0.137} \\
{\texttt{mBERT}}    & {14.640} & {14.242} & {5.437} & {1.006}  \\
{\texttt{IndoNLU}} & {13.021} & {12.993} & {5.225} & {1.088}  \\
\hline
\end{tabular}
\caption{Classification task}
\end{subtable}
\hfill
\begin{subtable}[h]{0.49\textwidth}
\begin{tabular}{lrrrr}
\hline
\multirow{2}{*}{\textbf{Method}} & \multicolumn{2}{c}{\textbf{load}} & \multicolumn{2}{c}{\textbf{infer}} \\
    & \textbf{CPU} & \textbf{GPU} & \textbf{CPU} & \textbf{GPU} \\
    \hline
\multicolumn{5}{c}{\textbf{SeqLab}} \\
\hline
{\texttt{CRF}} & {0.016} & {--} & {0.103} & {--} \\
{\texttt{Bi-LSTM}} & {0.261} & {0.261} & {0.715} & {0.455} \\
{\texttt{CNN}} & {1.035} & {0.981} & {0.657} & {0.479} \\
{\texttt{mBERT}} & {9.062} & {8.334} & {111.023} & {136.165} \\
{\texttt{IndoNLU}} & {6.359} & {6.236} & {15.701} & {11.234} \\
\hline \\

\end{tabular}

\caption{Labeling task}
\end{subtable}
\label{model-load-infer}
\vspace{-20px}
\end{table}

On almost all neural model cases, prediction using GPU is faster than the CPU. With Transformer, performance on prediction with GPU is significantly better with around 2-6 seconds faster (Table~\ref{model-load-infer} with the detail on Table~\ref{model-infer}). It shows the need to use GPU on production when we want to use those models which becomes a new cost that has to be considered. An anomaly happened for BERT in Sequence Labeling where using GPU show a slower performance.

%The Bi-LSTM and CNN could yield better results with hyper-parameter tuning. As it would be count as an additional cost on training, we did not do that.

\section{Discussion}
In the context of platform-as-a-service, we as a company learnt that users like to train many times everyday as they add a handful of new data and see whether there is an immediate improvement from it. This is infeasible with such long training duration of the Transformer when the platform needs to serve a big number of users without inferring additional, often unwanted cost to them. Indeed, managing users expectation is also another way to address this, but it is not necessarily easy. Anecdotally, we also see that the vast majority of our users prefer the fast training and prediction, rather than the fraction of higher accuracy from the transformer models.

The large size of transformer models also makes it costly to store, especially when model versioning is wanted by the users. Moreover, it also causes model loading and reloading into the memory to take longer. Distilling the models would also require more resources~\cite{kim2019research}. 
In addition, transformer approach often requires hyper-parameter tuning~\cite{murray2019auto}, which would significantly increase the cost of training.
In contrast, hyper-parameter tuning in classical model is cheaper~\cite{thornton2013auto}. 
%For neural model, which has large parameters, doing it would significantly increase the cost of training.

% As long as these problems are still there, we see little incentive for the companies within the emerging market to move on from the classical approaches and fully embrace the benefits of the deep transformer models, especially when the training data is limited. However, we understand the potential increase in the performance of Transformer models when we are dealing with a different task and/or when the data is abundant (e.g., machine translation). Hence, further analysis on the trade-off between cost and performance, in this case, might result in a different business decision, that is, one where we see using a deep neural model to be the more feasible solution.
Knowing these constraints, the decision to whether choose an advanced model like Transformer or the more classical ones depends on the company's capability. %When they are on a tight budget with limited machine resources and a lack of human expertise,
On a tight budget, with limited machine and human resources,
using classical algorithms is often sufficient for industrial use-cases, for which the labeled data are often relatively small in size, instead of jumping to the more advanced techniques. However, if the budget and resources are sufficient, Transformer-based models show great results and has more potentials to be fine-tuned towards better performance.
% With this experiment, we hope we could give more insight for companies in having AI-related business decisions. 

We propose the following direction to decrease the cost of adopting Transformer models for AI-pivoting businesses. Transfer learning can reduce the training cost, rather than training the model from scratch. In addition, transfer learning requires less data to train. We observe that lots of our clients actually share similar sets of labels (and similar training data that comes with them) %both for text classification and NER, 
but we still need to store their own model separately, while intuitively they should be able to either ``share a single model'' to some extent or that the knowledge stored in each of their models can be ``transferred'' more effectively into the others.

\section{Related Work}
Surveys to compare and analyze various algorithms have been done, both for text classification~\cite{1904.08067,2004.03705} and sequence labeling~\cite{yadav-bethard-2018-survey} which shows each advantages and limitations. 
All benchmarks gave good overview on what methods we could use, but the cost and resources were just briefly mention. Furthermore, the resource to host the model and its prediction time are not explored.

Knowing the long time and heavy resources it required, there are some studies for optimizing neural models. In terms of training duration, distributed training can be carried out to speed up the training process~\cite{dean2012large}. Further improvement can be achieved by compressing communications between GPUs or between servers through gradient quantization~\cite{1bitquant} or sparsification~\cite{dryden2016communication,aji2017sparse}. %CAMERA READY TAMBAH = han2015deep
However, these approaches can only be applied if the company or research institution has a lot of GPU resources to begin with.
  
Orthogonal to the training speed, 
%some research attempted to make the model more efficient in terms of size or prediction time. One 
a common way to make the model size smaller and faster for prediction without sacrificing its quality is to use knowledge distillation~\cite{hinton2015distilling}. 
To implement, a teacher system must be prepared, which is usually an ensemble of multiple larger neural models~\cite{kim2019research}. Preparing such teacher system requires additional resources. Therefore, the training cost for creating a lightweight model is usually higher.

\section{Conclusion}

% Our benchmark showed that the classic algorithms perform on par with deep neural methods in under-resourced data while being easier to implement and requiring significantly lower costs. 

Transformer-based pre-trained model outperform statistical methods on various NLP tasks.
However, it requires extra cost in terms of training time, memory to store the model, and its prediction time, compared to statistical approach. It also relies heavily on GPU which is still relatively uncommon and is expensive for cloud service. 
Our benchmark showed that the accuracy difference between the Transformer-based and statistical approach is 7\% at worst, therefore it is recommended for early adopters to use simple methods for their production environment.
After having enough resources to host large models, using pre-trained Transformers with the right (ideally distilled) base model should give the best results. We call for more research into efficient models to give more incentives to industry players in emerging market to use the Transformer from the get-go.

\bibliographystyle{splncs04}
\bibliography{mybibliography}

% \newpage
% \onecolumn
\appendix
\label{sec:appendix}

\section{Loading and Prediction Time}

A detail comparison for model loading time and prediction time for each method and dataset in CPU and GPU.

\begin{table}[ht]
\vspace{-20px}
\caption{\label{model-load}
Load time in seconds for CPU and GPU
}
\centering
\begin{tabular}{lrrrrrrrr}
\hline
\textbf{Method} & \textbf{CPU} & \textbf{GPU} & \textbf{CPU} & \textbf{GPU} & \textbf{CPU} & \textbf{GPU} & \textbf{CPU} & \textbf{GPU} \\
\hline
{} & \multicolumn{2}{c}{\textbf{Smalltalk}} & \multicolumn{2}{c}{\textbf{Healthcare}} & \multicolumn{2}{c}{\textbf{Telco}} & \multicolumn{2}{c}{\textbf{Sentiment}} \\
\hline
{\texttt{LR}} & {0.015} & {--} & {0.034} & {--} & {0.009} & {--} & {0.044} & {--} \\
{\texttt{SVM}} & {0.017} & {--} & {0.035} & {--} & {0.011} & {--} & {0.044} & {--} \\
{\texttt{Bi-LSTM}} & {1.022} & {1.169} & {7.363} & {7.338} & {1.245} & {1.243} & {6.220} & {6.425} \\
{\texttt{CNN}} & {1.510} & {1.368} & {8.029} & {8.020} & {1.639} & {1.657} & {7.651} & {7.408} \\
{\texttt{mBERT}} & {9.195} & {9.121} & {21.371} & {21.031} & {13.354} & {12.574} & \multicolumn{2}{c}{NA} \\
{\texttt{IndoNLU}} & {7.607} & {7.643} & {21.34} & {21.23} & {8.847} & {8.829} & {14.29} & {14.27} \\
\hline
{} & \multicolumn{2}{c}{\textbf{EntK}} & \multicolumn{2}{c}{\textbf{POS}} & \multicolumn{2}{c}{\textbf{TermA}} & \multicolumn{2}{c}{\textbf{Prod}} \\
\hline
{\texttt{CRF}} & {0.014} & {--} & {0.013} & {--} & {0.007} & {--} & {0.030} & {--} \\
{\texttt{Bi-LSTM}} & {0.409} & {0.416} & {0.403} & {0.407} & {0.131} & {0.123} & {0.100} & {0.099} \\
{\texttt{CNN}} & {1.531} & {1.433} & {1.931} & {1.868} & {0.478} & {0.456} & {0.199} & {0.167} \\
{\texttt{mBERT}} & {9.175} & {8.253} & {9.117} & {8.213} & {8.942} & {8.656} & {9.013} & {8.214} \\
{\texttt{IndoNLU}} & {6.420} & {6.265} & {6.311} & {6.197} & {6.360} & {6.251} & {6.346} & {6.229} \\
\hline
\end{tabular}
\vspace{-20px}
\end{table}

\begin{table}[ht]
\vspace{-20px}
\caption{\label{model-infer}
Prediction time in seconds for CPU and GPU
}
\centering
\begin{tabular}{lrrrrrrrr}
\hline
\textbf{Method} & \textbf{CPU} & \textbf{GPU} & \textbf{CPU} & \textbf{GPU} & \textbf{CPU} & \textbf{GPU} & \textbf{CPU} & \textbf{GPU} \\
\hline
{} & \multicolumn{2}{c}{\textbf{Smalltalk}} & \multicolumn{2}{c}{\textbf{Healthcare}} & \multicolumn{2}{c}{\textbf{Telco}} & \multicolumn{2}{c}{\textbf{Sentiment}} \\
\hline
{\texttt{LR}} & {0.369} & {--} & {0.395} & {--} & {1.344} & {--} & {0.214} & {--} \\
{\texttt{SVM}} & {0.101} & {--} & {0.179} & {--} & {0.121} & {--} & {0.212} & {--} \\
{\texttt{Bi-LSTM}} & {0.099} & {0.101} & {0.164} & {0.122} & {0.126} & {0.112} & {0.276} & {0.155} \\
{\texttt{CNN}} & {0.120} & {0.107} & {0.159} & {0.128} & {0.148} & {0.125} & {0.187} & {0.188} \\
{\texttt{mBERT}} & {4.470} & {0.901} & {6.442} & {1.071} & {5.400} & {1.045} & \multicolumn{2}{c}{NA} \\
{\texttt{IndoNLU}} & {3.425} & {0.963} & {5.680} & {1.147} & {4.417} & {1.129} & {7.377} & {1.114} \\
\hline
{} & \multicolumn{2}{c}{\textbf{EntK}} & \multicolumn{2}{c}{\textbf{POS}} & \multicolumn{2}{c}{\textbf{TermA}} & \multicolumn{2}{c}{\textbf{Prod}} \\
\hline
{\texttt{CRF}} & {0.063} & {--} & {0.115} & {--} & {0.039} & {--} & {0.195} & {--} \\
{\texttt{Bi-LSTM}} & {0.604} & {0.444} & {0.916} & {0.494} & {0.671} & {0.443} & {0.670} & {0.439} \\
{\texttt{CNN}} & {0.539} & {0.373} & {0.958} & {0.666} & {0.612} & {0.456} & {0.520} & {0.422} \\
{\texttt{mBERT}} & {109.581} & {134.869} & {112.975} & {134.451} & {111.513} & {134.653} & {110.023} & {140.688} \\
{\texttt{IndoNLU}} & {14.532} & {12.146} & {18.508} & {12.649} & {15.843} & {10.668} & {13.920} & {9.474}
 \\
\hline
\end{tabular}
\vspace{-20px}
\end{table}

\end{document}